\documentclass{article}

\usepackage{PRIMEarxiv}

\usepackage[utf8]{inputenc} 
\usepackage[T1]{fontenc}    
\usepackage{url}            
\usepackage{booktabs}       
\usepackage{amsfonts}       
\usepackage{nicefrac}       
\usepackage{microtype}      
\usepackage{lipsum}
\usepackage{fancyhdr}       
\usepackage{graphicx}       
\usepackage[colorlinks,urlcolor=blue,driverfallback=dvipdfm]{hyperref}
\graphicspath{{media/}}     

\usepackage{multirow}
\usepackage{makecell}
\usepackage{amsmath}
\newcommand{\eg}{\textit{e}.\textit{g}.}
\newcommand{\myPara}[1]{\vspace{.05in}\noindent\textbf{#1}}

\title{Deepfake Video Detection with Spatiotemporal Dropout Transformer
}

\author{
  Daichi Zhang$^{1,2}$, Fanzhao Lin$^{1,2}$, Yingying Hua$^{1,2}$, Pengju Wang$^{1,2}$, Dan Zeng$^{3}$ and Shiming Ge$^{1,2}$\thanks{Corresponding author.}\\
  $^1$ Institute of Information Engineering, Chinese Academy of Science\\
  $^2$ School of Cyber Security, University of Chinese Academy of Sciences\\
  $^3$ School of Communication and Information Engineering, Shanghai University\\
  \texttt{\{zhangdaichi,linfanzhao,huayingying,wangpengju,geshiming\}@iie.ac.cn} \\
  \texttt{\{dzeng\}@shu.edu.cn} \\
}

\begin{document}
\maketitle

\begin{abstract}
    While the abuse of deepfake technology has caused serious concerns recently, how to detect deepfake videos is still a challenge due to the high photo-realistic synthesis of each frame. Existing image-level approaches often focus on single frame and ignore the spatiotemporal cues hidden in deepfake videos, resulting in poor generalization and robustness. 
    The key of a video-level detector is to fully exploit the spatiotemporal inconsistency distributed in local facial regions across different frames in deepfake videos.
    Inspired by that, this paper proposes a simple yet effective patch-level approach to facilitate deepfake video detection via spatiotemporal dropout transformer. The approach reorganizes each input video into \emph{bag of patches} that is then fed into a vision transformer to achieve robust representation. 
    Specifically, a spatiotemporal dropout operation is proposed to fully explore patch-level spatiotemporal cues and serve as effective data augmentation to further enhance model's robustness and generalization ability. The operation is flexible and can be easily plugged into existing vision transformers.
    Extensive experiments demonstrate the effectiveness of our approach against 25 state-of-the-arts with impressive robustness, generalizability, and representation ability. 
\end{abstract}


\section{Introduction}
Deepfake~\cite{DBLP:journals/tog/SuwajanakornSK17} often refers to the technique that generates the images and videos by swapping the faces of source and target persons~\cite{DBLP:journals/corr/abs-1912-13457}, manipulating the original face attributes~\cite{lu2017attribute-guided}, or synthesizing an entire face that does not exist~\cite{thies2019deferred}. With the rapid development of face generation and manipulation methods, especially after generative adversarial networks (GANs) were proposed \cite{goodfellow2014generative}, deepfake videos can be easily produced by accessible online tools, such as FaceSwap\footnote{https://github.com/MarekKowalski/FaceSwap/} and Deepfakes\footnote{https://github.com/deepfakes/faceswap}, but can barely be distinguished by the human eyes, leading to significant threaten to the public social, cyber and even political security, such as fabricating evidence and ruining political discourse~\cite{DBLP:journals/tog/SuwajanakornSK17}. Thus, it is very critical to develop effective solutions to detect deepfake videos.

\begin{figure}[t]
    \centering
    \includegraphics[width=0.9\linewidth]{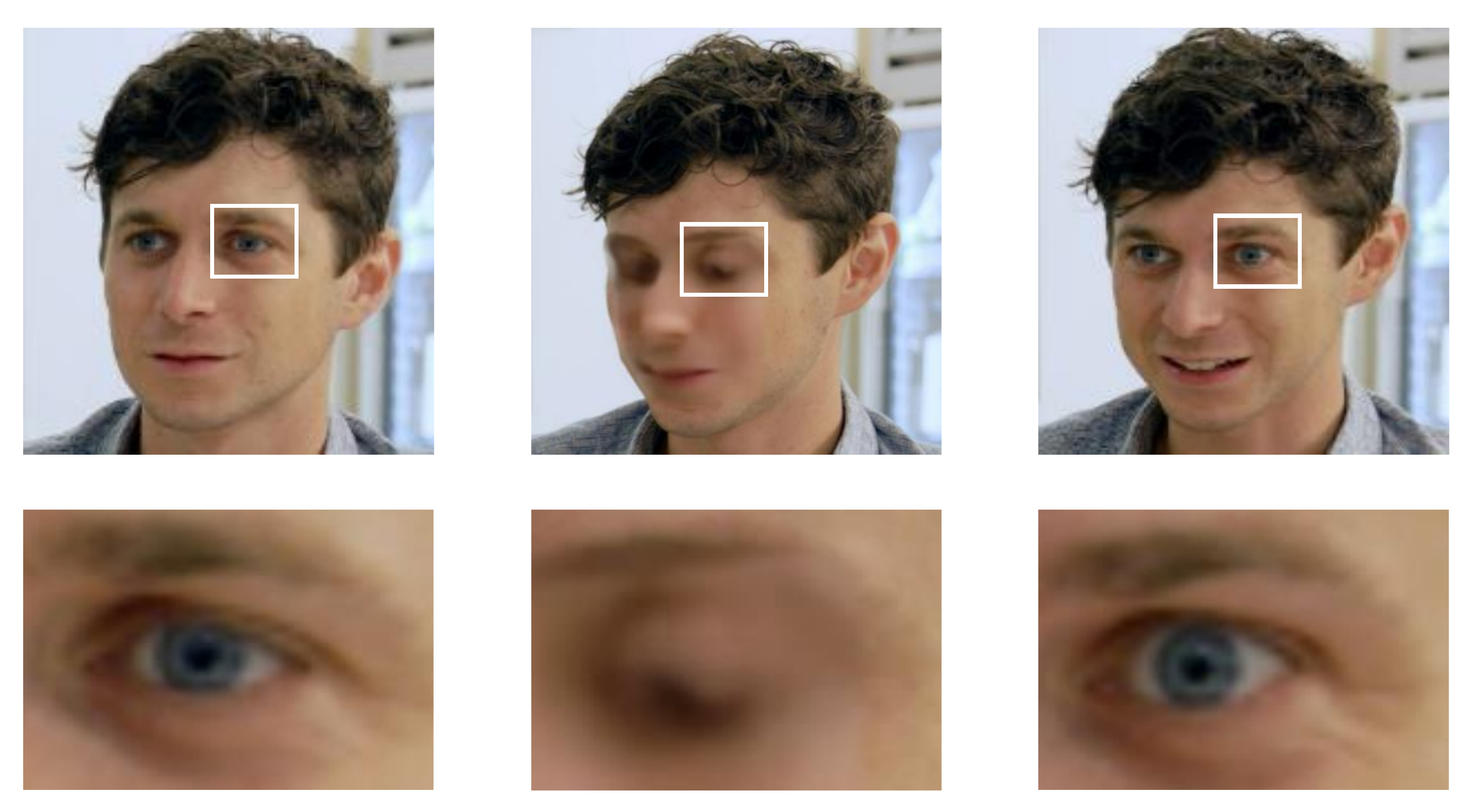}
    \caption{The detection of deepfake videos is challenging due to high photo-realistic frame synthesis. Thus, our approach leverages patch-level spatiotemporal inconsistency in facial regions across frames to facilitate deepfake video detection.}\label{fg:motivation}
\end{figure}

Existing approaches usually formulate the deepfake video detection task as a binary classification problem and can be divided into two major categories: image-level and video-level approaches. Image-level approaches perform frame-wise detection by mining the pattern difference between real and fake images~\cite{DBLP:conf/iccp/PanZL12,DBLP:conf/sswmc/0002SS07,Rssler2019FaceForensics,li2020face}.
Generally, these approaches can well exploit the spatial cues in the frames but neglect the temporal cues in the video. Thus, image-level approaches may be limited when detecting deepfake videos. Typically, advanced deepfake approaches may generate extremely genuine facial images without leaving spatial defect but they cannot properly avoid temporal inconsistency since deepfake videos are always generated frame-by-frame. Unlike image-level approaches, recent video-level approaches~\cite{guera2018deepfake,DBLP:conf/ijcai/ZhangLL0G21,DBLP:conf/ih/AmeriniC20} focus on the sequence patterns and aim to explore the spatiotemporal inconsistency to detect.
However, this spatiotemporal inconsistency is distributed dynamically in different local regions and frames, which is extremely difficult to be captured, as shown in Fig.~\ref{fg:motivation}, only the eye regions in different frames are visibly inconsistent. Existing video-level approaches are not specially designed for this inconsistency and can not properly capture the spatiotemporal cues which hide in deepfake videos, which makes these approaches poorly generalized and vulnerable, even cannot achieve comparable performance to image-level approaches. Therefore, a key of deepfake video detection approach is to effectively learn the discriminative representations to describe the spatiotemporal inconsistency.

Towards this end, this paper proposes an effective patch-level deepfake video detection framework, named spatiotemporal dropout transformer to capture the spatiotemporal inconsistency effectively. In the approach, the input video is first extracted into facial frame sequence then each frame is grid-wisely cropped into non-overlapping facial patches, which are subsequently reorganized into \emph{bag of patches} and fed into a vision transformer to learn the discriminative representations describing the dynamical spatiotemporal cues in local facial regions across different frames as well as achieving robust representation capacity.

Specifically, a spatiotemporal dropout~(STD) operation is designed to fully explore the spatiotemporal inconsistency at patch-level. The STD operation performs temporal dropout and spatial dropout step-wisely. During the temporal dropout, we randomly drop part of the extracted frames after we obtain the frame sequence from input video. Then during the spatial dropout, we randomly drop part of facial patches grid-wisely cropped from each remaining frame and reorganize the remaining patches as a \emph{bag of patches} to train a vision transformer. The bag of patches still contains all facial regions in the original face to preserve the whole information but the data to be processed is largely reduced since the dropout operation. In this way, the inconsistency distributed in local facial patches across different frames in deepfake video is fully represented and explored. And since the dropout operation is random, massive different bag of patches instances can be generated from the same input video in different training iterations, which also serves as data augmentation for more generalized and robust detection. Moreover, the STD operation is flexible and could be plugged into existing vision transformers~(ViTs).

Our main contributions can be summarized as three folds. First, we propose an effective patch-level deepfake video detection framework, Spatiotemporal Dropout Transformer. In our approach, input videos are reorganized as \emph{bag of patches} instances which are then fed into a vision transformer to achieve strong representation capacity. Second, we design a simple yet effective spatiotemporal dropout operation, which can fully explore the patch-level spatiotemporal inconsistency hidden in deepfake videos and also serve as an effective data augmentation to further improve the model's robustness and generalization ability. Besides, our STD operation can be plugged into existing ViTs. Third, we conduct extensive experiments on three public benchmarks to demonstrate that our approach outperforms 25 state-of-the-arts with impressive robustness, generalizablity, and representation ability.

\section{Related Work}
\myPara{Deepfake Generation.}~Deepfake refers to the techniques of synthesizing or manipulating human face images or videos\cite{DBLP:conf/icpr/TolosanaRFV20}. Early deepfake is generated by hand-crafted features designed by researchers such as face landmarks, and some post-processing methods are utilized to make the generating artifacts invisible. For example, \cite{DBLP:conf/cvpr/0001VRTPT14} designs a face reenactment system based on face matching and \cite{DBLP:journals/tog/DaleSJVMP11} proposes a 3D multilinear model to track the face movement in video and minimize the blending boundaries. Since these traditional generation methods often suffer from generating artifacts and visual quality, deep learning-based deepfake generation methods are developed for more realistic face synthesis. For example, generative adversarial networks~(GANs)~\cite{goodfellow2014generative} have enabled plenty of high-quality face manipulation~\cite{thies2016face2face:,Li_2020_CVPR} and face synthesis~\cite{DBLP:conf/cvpr/KarrasLA19,DBLP:conf/cvpr/ChoiCKH0C18,DBLP:conf/cvpr/KarrasLAHLA20}.
Although these methods can achieve high-quality generation results, they work in a frame-by-frame way to generate deepfake videos, and the spatiotemporal cues are difficult to eliminate, which can serve as an effective discriminative clue.

\myPara{Deepfake Detection.}~Since deepfake has brought severe threats to society, a variety of deepfake detection approaches have been proposed. Early detection approaches focus on hand-crafted features which are limited at that time, such as~\cite{DBLP:conf/iccp/PanZL12,DBLP:conf/sswmc/0002SS07} utilize image statistic features to detect. With the development of deep learning, researchers begin to utilize DNNs to perform deepfake detection, which can be further divided into two categories: image-level approaches and video-level approaches. 
Image-level approaches focus on extracting discriminative image-level features for deepfake detection, such as~\cite{Rssler2019FaceForensics} utilizes the XceptionNet to detect deepfakes and~\cite{li2020face} detects the blending boundary of two face images. All these approaches can achieve impressive performance on image-level deepfake detection but ignore the temporal cues hidden in deepfake videos, resulting in poor performance when detecting deepfake videos.
Video-level approaches pay more attention to the sequence feature and many general video models have been applied to detect deepfake videos, such as~\cite{DBLP:conf/ijcai/ZhangLL0G21,guera2018deepfake,DBLP:conf/ih/AmeriniC20}. However, these models usually are not specially designed for deepfake video detection task and are not capable to properly capture the spatiotemporal inconsistency dynamically distributed in local facial regions across different frames. For example, TD-3DCNN~\cite{DBLP:conf/ijcai/ZhangLL0G21} only consider the inter-frame inconsistency in frame level while ignoring the intra-frame inconsistency cues in spatial domain.

\myPara{Vision Transformer.}~Transformers~\cite{DBLP:conf/nips/VaswaniSPUJGKP17} have achieved impressive performance in natural language processing~(NLP) tasks due to their strong representation capacity, such as BERT~\cite{DBLP:conf/naacl/DevlinCLT19}. Recently, researchers have proved that transformers can also achieve excellent performance on a variety of computer vision tasks. Specifically, vision transformer (ViT)~\cite{DBLP:conf/iclr/DosovitskiyB0WZ21} utilizes the same self-attention mechanism and crops an image into a sequence of flattened patches as the input token sequence used in the NLP task to train the transformer encoder for different downstream tasks, such as classification and object detection~\cite{DBLP:conf/iclr/DosovitskiyB0WZ21,carion2020end}. Various ViT architectures have been proposed~\cite{DBLP:journals/corr/abs-2103-14030,DBLP:journals/corr/abs-2103-15691,DBLP:journals/corr/abs-2102-12122} recently and related experiments results further demonstrate ViT also has remarkable representation capacity when dealing with images and videos~\cite{han2020survey}. Many previous video-level detectors choose traditional CNN as their backbones, which may restrict their performance since CNN's limited representation capacity compared to transformer, such as~\cite{DBLP:conf/ijcai/ZhangLL0G21}. To enhance the model's representation capacity in deepfake video detection, more powerful backbones are needed. There are also some existing works that apply ViT to deepfake detection tasks, such as~\cite{DBLP:journals/corr/abs-2102-11126,DBLP:journals/corr/abs-2107-02612,DBLP:journals/corr/abs-2104-01353}, but these approaches focus more on designing ViT architecture or combining with other approaches without exploring the intrinsic characteristics of deepfake video detection task, such as the spatiotemporal cues. Therefore, it is necessary to design an effective and flexible framework to incorporate ViT in deepfake video detection which can make full use of the dynamical cues distributed in local regions across frames (\eg, patch-level spatiotemporal cues in deepfake videos) to further improve model performance.

\begin{figure}[t]
   \centering
   \includegraphics[width=\linewidth]{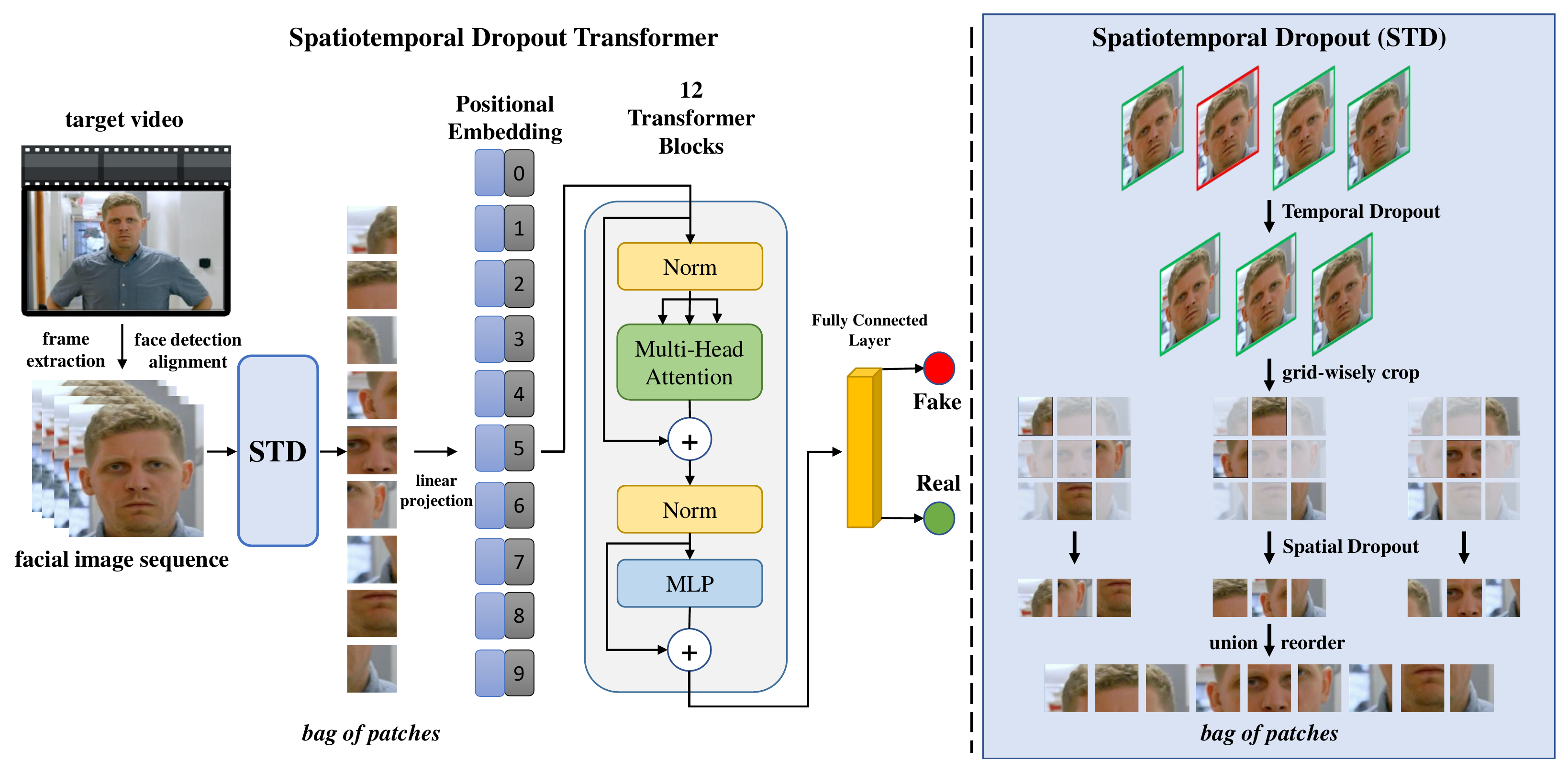}
   \caption{\label{figure:framework} The framework of Spatiotemporal Dropout Transformer for deepfake video detection. We first employ frame extraction, face detection and alignment to get facial image sequence $\mathcal{F}=\{\mathbf{x}_i\}_{i=1}^{n}$. Then each facial image $\mathbf{x}_i \in \mathcal{F}$ is processed by STD operation to get facial patch set ${P}_{i^{'}} = \{\mathbf{p}_{i^{'},j}\}$ and further reorganized as a \emph{bag of patches} instance $\mathcal{P}$ which is then fed into our vision transformer backbone to learn the discriminative representation and detect.}
\end{figure}

\section{The Approach}

\subsection{Problem Formulation}\label{sec:3.1}

In our approach, the objective of deepfake video detection is to learn a discriminative binary classifier $\phi$ to identify a video clip consisting of $n$ frames $\mathcal{V}=\{\mathbf{f}_i\}_{i=1}^{n}$ into real or fake. 
Thus, learning $\phi$ can be formulated as an energy minimization problem that can be solved by:
\begin{equation}\label{eq:problem}
    \mathbb{W}^{*} = \arg\min_{\mathbb{W}}\sum_{\mathcal{V}\in\mathcal{D}}{\mathbb{E}(\phi(\mathbb{W};\mathcal{V}),l)},
\end{equation}
where $\mathbb{W}^{*}$ is the learned optimal output of the detector parameters, $\mathcal{D}$ is the training set, $l\in\{0,1\}$ is the video label and $\mathbb{E}$ is an energy function to measure detection loss.

Considering that the spatiotemporal inconsistency existing in deepfake videos is dynamically distributed in different local facial regions across different frames, a key to be addressed by the detector is providing a flexible and exhaustive way to fully explore the spatiotemporal cues and aggregate them to form and learn a typical discriminative features to detect. To achieve that, the detector $\phi$ should be able to accept dynamical input instances containing spatiotemporal cues to learn a powerful backbone with stronger representation capacity. Towards this end, we take a vision transformer as the backbone and incorporate a simple yet effective Spatiotemporal Dropout operation to fully explore the dynamical spatiotemporal inconsistency across different frames at patch-level, as presented in Fig.~\ref{figure:framework} and introduced in the following.

\subsection{Spatiotemporal Dropout}
The Spatiotemporal Dropout operation aims to generate the dynamical input instances from input videos, which contain spatiotemporal inconsistency cues distributed in different facial regions across different frames. Therefore, we perform it in an efficient step-wise manner, including \emph{Temporal Dropout} and \emph{Spatial Dropout}.

\myPara{Temporal Dropout.}~
For each video, we first randomly sample $n$ consistent raw frames $\{\mathbf{f}_i\}_{i=1}^{n}$ and employ face detection and alignment to each frame $\mathbf{f}_i$ to get a facial image sequence $\mathcal{F}=\{\mathbf{x}_i\}_{i=1}^{n}$. Then we randomly discard part of facial images following a uniform distribution, with defined temporal dropout rate $\alpha$ and remaining $(1 - \alpha) \times n$ facial images, which can be formulated as:
\begin{equation}\label{eq:TD}
    \begin{aligned}
        \mathcal{F}_{T} =&~ \subseteq(\mathcal{F}, \alpha)\\
        =&~ \subseteq(\{\mathbf{x}_1, \mathbf{x}_2, ..., \mathbf{x}_{n} \}, \alpha)\\
        =&~ \{\mathbf{x}_{k_1}, \mathbf{x}_{k_1 + 1}, ..., \mathbf{x}_{k_1 + (1 - \alpha) * n - 1}\},\\
        &~ k_1 \sim \mathcal{U}(1, \alpha \times n + 1), k_1 \in \mathbb{Z}, \alpha \in (0, 1)
    \end{aligned}
\end{equation}
where $\subseteq$ means the subset operation, $\alpha$ is the temporal dropout rate, $k_1$ is a random start index following a uniform distribution, and $\mathcal{F}$ is the preprocessed facial image sequence. The uniform distribution can prevent center bias and we finally get a sparse sequence $\mathcal{F}_{T}$.

\myPara{Spatial Dropout.}~
For each remaining facial image $\mathbf{x}_{i^{'}} \in \mathcal{F}_{T}$, we grid-wisely crop $\mathbf{x}_{i^{'}}$ into $m$ regular non-overlapping facial patches ${P}_{i^{'}}=\{\mathbf{p}_{i^{'},j}\}_{j=1}^{m}$,
then randomly discard part of patches following a uniform distribution, with defined spatial dropout rate $\beta$ and remaining $(1 - \beta) \times m$ facial patches, which can be formulated as:
\begin{equation}\label{eq:SD}
    \begin{aligned}
        {P}_{i^{'},S} =&~ \subseteq({P}_{i^{'}}, \beta)\\
        =&~ \subseteq(\{\mathbf{p}_{i^{'},1}, \mathbf{p}_{i^{'},2}, ..., \mathbf{p}_{i^{'},m} \}, \beta)\\
        =&~ \{ \mathbf{p}_{i^{'}, k_2}, \mathbf{p}_{i^{'}, k_2 + 1}, ..., \mathbf{p}_{i^{'}, k_2 + (1 - \beta) * m - 1}\},\\
        &~ i^{'} \in [k_1, k_1 + (1-\alpha) \times n - 1],\\
        &~ k_2 \sim \mathcal{U}(1, \beta \times m + 1), k_2 \in \mathbb{Z}, \beta \in (0, 1)
    \end{aligned}
\end{equation}
where $\subseteq$ means the subset operation, $\beta$ is the spatial dropout rate, $k_2$ is a random start index following a uniform distribution and $P_{i^{'}}$ is the cropped facial patch set of $\mathbf{x}_{i^{'}}$. The uniform distribution can prevent center bias and we finally get a sparse patch set ${P}_{i^{'},S}$.

\myPara{Bag of Patches.}~
After we get all facial patch sets $\{ {P}_{i^{'},S} \}_{i^{'}=k_1}^{k_1 + (1-\alpha)*n - 1}$, we can generate our bag of patches instance $\mathcal{P}$ by first collecting all ${P}_{i^{'},S}$ together:
\begin{equation}\label{eq:BoP}
    \begin{aligned}
        \mathcal{P} = \cup \{ {P}_{i^{'},S} \}_{i^{'}=k_1}^{k_1 + (1-\alpha)*n - 1},
    \end{aligned}
\end{equation}
where $\cup$ means the union of sets. To guarantee each bag of patches contains all facial regions of original face, we ensure the index of each patch $\mathbf{p}_{i^{'},j} \in P_{i^{'},S}$ without repetition by controlling $k_2$ in Eq.(\ref{eq:SD}) and reorganize the patch order in $\mathcal{P}$ in original face patch order.

The whole algorithm of our spatiotemporal dropout can be described as follows:
\begin{itemize}
    \item Step 1: Extract and randomly sample $n$ consistent raw frames from input video $\mathcal{V}=\{\mathbf{f}_i\}_{i=1}^{n}$.
    \item Step 2: Employ face detection and alignment to each frame $\mathbf{f}_i$ to get facial image sequence $\mathcal{F}=\{\mathbf{x}_i\}_{i=1}^{n}$.
    \item Step 3: Randomly discard $\alpha \times n$ images from $\mathcal{F}$ through temporal dropout to get $\mathcal{F}_{T}$.
    \item Step 4: Grid-wisely crop each remaining facial image $\mathbf{x}_{i^{'}} \in \mathcal{F}_{T}$ into $m = row \times col$ patches to get facial patch set $P_{i^{'}}$.
    \item Step 5: Randomly discard $\beta \times m$ patches from each facial patch set $P_{i^{'}}$ to get $P_{i^{'},S}$.
    \item Step 6: Collect and reorder all facial patch set $P_{i^{'},S}$ to get one bag of patches instance $\mathcal{P}$.
\end{itemize}

Different from many existing video-level approaches which focus on single frames or the general 3D feature of video, our STD is specially designed to capture the patch-level spatiotemporal cues hidden in deepfake videos. By applying STD operation during training, the spatiotemporal inconsistency in facial regions across different frames is fully explored and learned by our ViT backbone. By introducing the dropout operation, the data to be processed is largely reduced. Besides, since the dropout operation is random, we can generate massive different bag of patches instances $\mathcal{P}$ from the same input video $\mathcal{V}$ in each different training iteration, exhaustively exploiting the dynamical spatiotemporal cues and also serving as data augmentation to further improve our model's robustness and generalization ability. Moreover, the STD operation is flexible and could be plugged into existing ViTs, which is further discussed in the Experiments section.

\subsection{Overall Architecture}
After obtaining massive bag of patches instances $\mathcal{P}$ through our STD operation above, we feed them into our ViT backbone to optimize the final classifier $\phi$ to perform detection, which turns Eq.(\ref{eq:problem}) into following formulation:
\begin{equation}\label{eq:problem1}
    \mathbb{W}^{*} = \arg\min_{\mathbb{W}}\sum_{\mathcal{V}\in\mathcal{D}}\sum_{\mathcal{P}\in{STD}(\mathcal{V})}{\mathbb{E}(\phi(\mathbb{W};\mathcal{P}),l)},
\end{equation}
where $\mathbb{W}^{*}$ is the learned optimal output of the detector parameters, $\mathcal{V}$ is the video clip in training set $D$, $l\in\{0,1\}$ is the video label and $\mathbb{E}$ is an energy function to measure detection loss.

The overall architecture of our proposed approaches is presented in Fig.~\ref{figure:framework}. During the training process, we first employ frame extraction, face detection, and alignment to input video $\mathcal{V}$ to get facial image sequence $\mathcal{F}=\{\mathbf{x}_i\}_{i=1}^{n}$. Then the facial image sequence $\mathcal{F}$ is processed by our STD operation to generate bag of patches instance $\mathcal{P}$, which is then linearly projected with positional embedding and fed into our transformer encoder to achieve a stronger representation capacity. Since the dropout operation is random, massive different bag of patches instances are generated in different training iterations from the same input video, exhaustively mining the dynamic spatiotemporal cues at patch-level and also serving as data augmentation. The representations learned by the transformer encoder are input to a fully connected layer and output the prediction of being fake or real. For our ViT backbone, we choose the most basic ViT-Base-16 model presented in~\cite{DBLP:conf/iclr/DosovitskiyB0WZ21} which contains 12 transformer blocks consisting of two normalization layers, one multi-head attention block, and one MLP head. A $Binary~Cross~Entropy$ loss function is employed as our criteria and energy function in Eq.(\ref{eq:problem1}). During inference, test videos are processed with the same procedure in training and the model would output the prediction of being fake or real.

\section{Experiments}
We conduct experiments and present a systematic analysis to demonstrate the effectiveness of our proposed Spatiotemporal Dropout Transformer~(STDT). First, we make comparisons with 25 state-of-the-arts on three popular benchmarks. Then, we conduct visualizations and experiments under different perturbations and across different datasets to demonstrate its robustness, cross-dataset generalization and representation ability. Finally, a series of ablation analysis are performed to investigate the impact of each key components of our approach.

\subsection{Experimental Settings}
\myPara{Datasets}.~We evaluate our model on FaceForensics++~(FF++)~\cite{Rssler2019FaceForensics}, DFDC~\cite{dolhansky2019the} and Celeb-DF(v2)~\cite{DBLP:conf/cvpr/LiYSQL20} datasets. 
FF++ contains 1,000 original videos and corresponding 5$\times$1,000 manipulated videos by using five different generation methods~(including Deepfakes, Face2Face, FaceShifter, FaceSwap, and NeuralTextures) with different compression rates~(raw, c23, and c40 from no to high compression), where we choose the raw data and select 1,600 for training, 200 for validation and 200 for testing for each subset.
DFDC (the Deepfake Detection Challenge dataset) contains original videos recorded from 430 hired actors and over 400G fake videos are synthesized by several deepfake generation methods. Among all videos, we randomly select 6,261 for training, 800 validation, and 781 for testing.
Celeb-DF(v2) contains 590 original videos covering different ages, genders, and ethnics, and 5,639 corresponding synthesized videos (4,807 for training, 1,203 for validation, and 518 for testing).

\myPara{Implementation Details.}~We use FFmpeg\footnote{https://ffmpeg.org/} to extract all frames of original videos and choose a pre-trained MobileNet\footnote{https://github.com/yeephycho/tensorflow-face-detection} as the face detector for all datasets. The extracted face images are then aligned and resized into $384\times384$ shape. 
For our vision transformer backbone, we choose the most basic ViT-Base-16 model described in~\cite{DBLP:conf/iclr/DosovitskiyB0WZ21} as our backbone. Specifically, our ViT-Base-16 backbone contains 12 layers with the dropout rate set to 0.1 and 12 self-attention heads with the attention dropout rate set to 0.0. And we choose $Binary~Cross~Entropy~(BCE)$ as our loss and energy function in Eq.(\ref{eq:problem1}). The embedded vector dimension of projected flatten tokens is 768 and the dimension of MLP header is 3072. Our ViT was pretrained on ImageNet and we adopt SGD optimizer and OneCylcleLR strategy. The global learning rate is set to $10^{-3}$ and the weight decay is set to $10^{-4}$. We set the raw frame sequence length $n$ to 24, temporal dropout rate $\alpha$ to $1/4$, spatial dropout rate $\beta$ to $17/18$ and each face is grid-wisely cropped into $6\times6$ patches. Furthermore, no additional augmentation methods are employed during training for a fair evaluation of our STD operation.

\subsection{State-of-the-Art Comparison}\label{sec:4.1}
\begin{table}[t]
   \centering
   \caption{AUC (\%) comparisons with 25 state-of-the-art approaches on three popular benchmarks.}\label{tab:tab1}
       \begin{tabular}{ccccc}
          \toprule
          \multirow{1}{*}{Approach} & FF++ & DFDC & Celeb-DF & Year\\
          \midrule
          Two-stream~\cite{DBLP:conf/cvpr/ZhouHMD17}  & 70.7 & 61.4 & 53.8 & 2017\\
          MesoNet~\cite{DBLP:conf/wifs/AfcharNYE18} & 84.7 & 75.3 & 54.8 & 2018\\
          Head-Pose~\cite{yang2019exposing} & - & 55.9 & 54.6 & 2019\\
          Vis-Art~\cite{8638330} & 78.0 & 66.2 & 55.1 & 2019\\
          Multi-Task~\cite{nguyen2019multi} & 76.3 & 53.6 & 54.3 & 2019\\
          Warp-Art~\cite{DBLP:conf/cvpr/LiL19c} & 93.0 & 75.5 & 64.6 & 2019\\
          XceptionNet~\cite{Rssler2019FaceForensics} & 95.5 & 69.7 & 65.5 & 2019\\
          CapsuleNet~\cite{DBLP:journals/corr/abs-1910-12467} & 96.6 & 53.3 & 57.5 & 2019\\
          CNN-RNN~\cite{sabir2019recurrent} & 80.8 & 68.9 & 69.8 & 2019\\
          CNN-Spot~\cite{DBLP:conf/cvpr/WangW0OE20} & 65.7 & 72.1 & 75.6 & 2020\\
          X-Ray~\cite{li2020face} & 92.8 & 65.5 & 79.5 & 2020\\
          TwoBranch~\cite{DBLP:conf/eccv/MasiKMGA20} & 93.2 & - & 76.6 & 2020\\
          PatchBased~\cite{DBLP:conf/eccv/ChaiBLI20} & 57.8 & 65.6 & 69.9 & 2020\\
          AudioVis~\cite{mittal2020emotions} & - & 84.4 & - & 2020\\
          TD-3DCNN~\cite{DBLP:conf/ijcai/ZhangLL0G21} & 72.2 & 79.0 & 88.8 & 2021\\
          MAT~\cite{zhao2021multi} & 97.6 & - & - & 2021\\
          Lips~\cite{DBLP:conf/cvpr/HaliassosVPP21} & 97.1 & 73.5 & 82.4 & 2021\\
          DIANet~\cite{DBLP:conf/ijcai/HuXWLW021} & 90.4 & 90.5 & 70.4 & 2021\\
          SPSL~\cite{liu2021spatial} & 96.9 & 66.2 & - & 2021\\
          FD$^{2}$Net~\cite{DBLP:conf/cvpr/Zhu0FLL21} & 99.5 & 66.1 & - & 2021\\
          ConvViT~\cite{DBLP:journals/corr/abs-2102-11126} & - & 91.0 & - & 2021\\
          EffViT~\cite{DBLP:journals/corr/abs-2107-02612} & - & 91.9 & - & 2021\\
          DistViT~\cite{DBLP:journals/corr/abs-2104-01353} & - & 97.8 & - & 2021\\
          VFD~\cite{DBLP:journals/corr/abs-2203-02195} & - & 98.5 & - & 2022\\
          ICT~\cite{DBLP:journals/corr/abs-2203-01318} & 98.6 & - & 94.4 & 2022\\
          \midrule
          \textbf{STDT~(Ours)} & \textbf{99.8} & \textbf{99.1} & \textbf{97.2} & 2022\\
          \bottomrule
       \end{tabular}
\end{table}

\begin{table}[t]
   \centering
   \caption{Intra-datasets results on Celeb-DF(v2), DFDC and five subsets of FaceForensics++ (Deepfakes, Face2Face, FaceShifter, FaceSwap and NeuralTextures).
   }\label{tab:tab2}
       \begin{tabular}{ccccccc}
          \toprule
              {Datasets} & {ACC(\%)} & {AUC(\%)} & {REC(\%)} & {PRE(\%)} & {F1(\%)} \\
              \midrule
              {Celeb-DF} & 91.70 & 97.21 & 95.01 & 92.55 & 93.76\\
              {DFDC} & 97.44 & 99.14 & 98.63 & 98.33 & 98.48\\
              {Deepfakes} & 97.97 & 99.76 & 95.70 & 100.0 & 97.80\\
              {Face2Face} & 98.01 & 99.09 & 97.94 & 97.94 & 97.94\\
              {FaceShifter} & 98.64 & 99.88 & 98.28 & 99.13 & 98.70\\
              {FaceSwap} & 98.61 & 99.89 & 99.11 & 98.23 & 98.67\\
              {NeuralTextures} & 91.88 & 97.98 & 97.98 & 97.50 & 97.74\\
          \bottomrule
       \end{tabular}
\end{table}

To demonstrate the advantage of our approach, we make comparisons with 25 state-of-the-art approaches on Celeb-DF(v2), DFDC and FaceForensics++ Deepfakes subset. These approaches include image-level detectors which use the average frame-level scores as the final prediction, video-level detectors such as TD-3DCNN~\cite{DBLP:conf/ijcai/ZhangLL0G21} and RNN~\cite{sabir2019recurrent}, and recent approaches based on ViT~\cite{DBLP:journals/corr/abs-2102-11126,DBLP:journals/corr/abs-2107-02612,DBLP:journals/corr/abs-2104-01353}. Training and testing on the same datasets aim to explore the model’s capacity of capturing the deepfake cues in deepfake videos. The area under the curve~(AUC) score is used as the metric and the results are presented in Tab.~\ref{tab:tab1}.

From the table, we can easily observe that our STDT consistently achieves the highest AUC score on all three datasets, ie, 99.8\% on FaceForensics++, 99.1\% on DFDC and 97.2\% on Celeb-DF(v2), which is at least 0.3\%, 0.6\% and 2.8\% higher than the state-of-the-arts, demonstrating the effectiveness of our proposed approach. Specifically, DFDC and Celeb-DF contain higher quality videos with a variety of scenarios, people groups and generation methods and the FF++ contains videos from the Internet, indicating our approach can still achieve impressive performance in a variety of situations. The main reason is that our approach focuses on capturing the patch-level inconsistency from both inter-frame and intra-frames rather than inter-frame inconsistency like~\cite{DBLP:conf/ijcai/ZhangLL0G21} .

\begin{figure}[t]
\tiny
    \centering
    \includegraphics[width=0.75\linewidth]{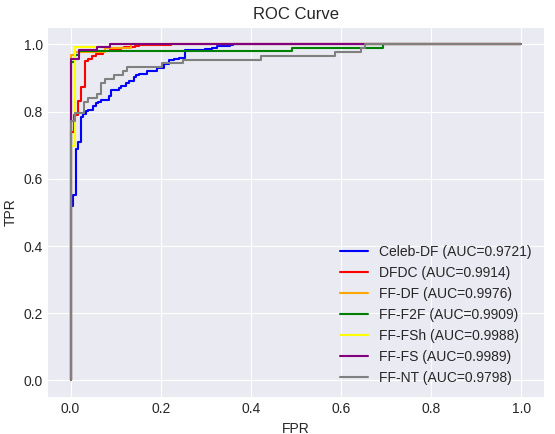}
    \caption{ROC curves on Celeb-DF(v2), DFDC and five subsets of FaceForensics++ datasets~(Deepfakes, Face2Face, FaceShifter, FaceSwap and NeuralTextures).}\label{figure:roc}
\end{figure}

Additionally, as the FF++ dataset contains several subsets generated by different deepfake methods, we conduct intra-dataset experiments on each subset respectively to further evaluate our proposed approach. We choose the accuracy~(ACC), the area under the curve~(AUC), recall~(REC), precision~(PRE), and F1 score as our evaluation metrics. The results are presented in Tab.~\ref{tab:tab2} together with Celeb-DF(v2) and DFDC. From the table, it can be observed that our proposed approach achieves impressive performance on all three datasets with all ACC higher than 90\%, AUC higher than 97\%, demonstrating the effectiveness and the ability of our approach on handling various deepfake generation methods. 
Especially on the Deepfakes, Face2Face, FaceShifter, and FaceSwap subset of FF++, we achieve all AUC higher than 99\%. The corresponding ROC curves on each dataset are also presented in Fig.~\ref{figure:roc} for better comprehension.
After analyzing some samples from these datasets, we find that all these different generation methods will result in subtle visual artifacts around some facial regions across frames, which may be disguised by the video quality or so subtle that could be easily ignored. These spatiotemporal cues are hard to be captured by human eyes, but the results demonstrate that our approach can effectively capture these inconsistencies and distinguish the deepfake videos generated by various methods with impressive performance.

\begin{table}[t]
   \centering
   \caption{Robustness experiments on Celeb-DF(v2) under five representative augmentation methods.}\label{tab:tab6}
       \begin{tabular}{ccccccc}
          \toprule
              {Augmentation} & {ACC(\%)} & {AUC(\%)} & {REC(\%)} & {PRE(\%)} & {F1(\%)} \\
              \midrule
              {Flip}& 90.54 & 96.56 & 95.01 & 90.99 & 92.95\\
              {Blur}& 89.01 & 94.69 & 92.94 & 90.54 & 91.73\\
              {Bright}& 87.84 & 94.13 & 92.06 & 89.68 & 90.86\\
              {Compress}& 89.38 & 95.74 & 96.76 & 88.20 & 92.29\\
              \makecell[c]{Gaussian Noise}& 87.45 & 94.25 & 96.76 & 85.90 & 91.01\\
              {Origin} & 91.70 & 97.21 & 95.01 & 92.55 & 93.76\\
          \bottomrule
       \end{tabular}
\end{table}

\begin{figure*}[t]
    \centering
    \includegraphics[width=\linewidth]{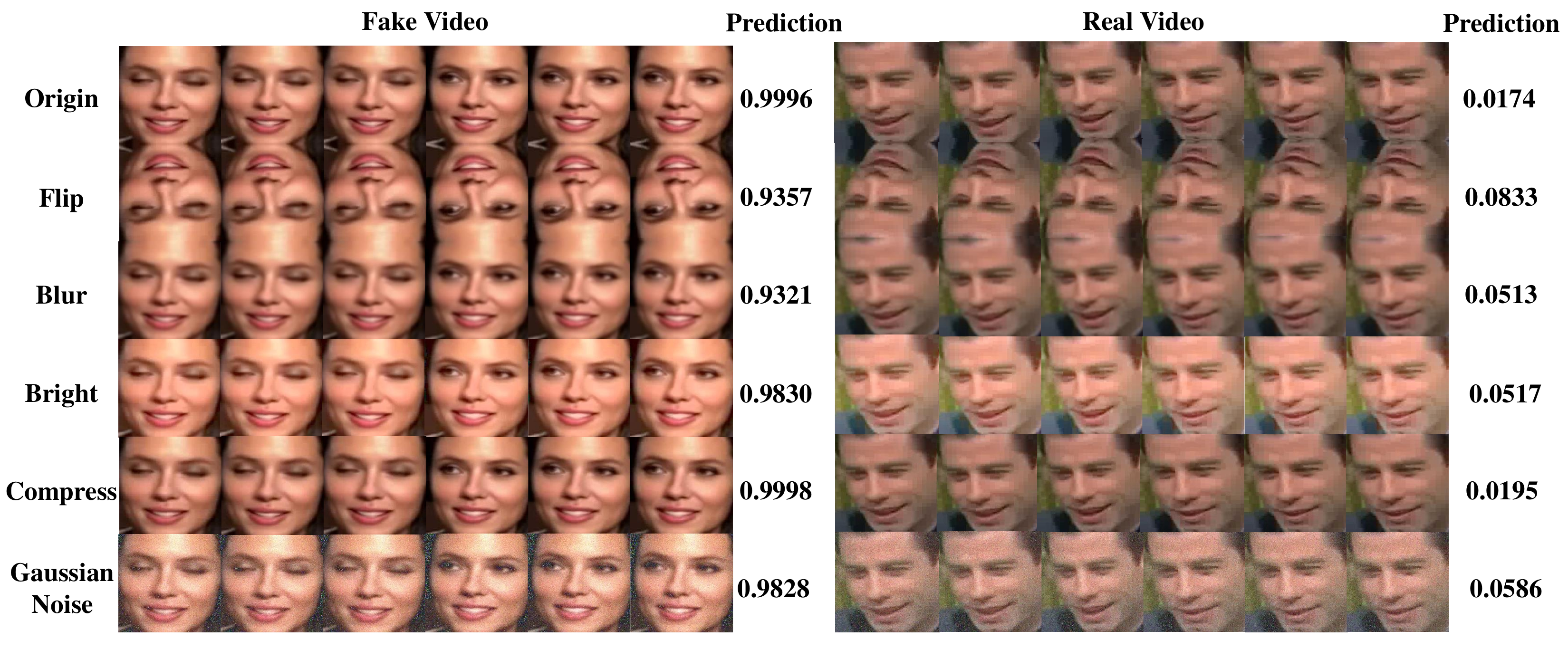}
    \caption{Two video examples from Celeb-DF(v2) with five augmentation methods and the predicted possibilities of being fake.}\label{figure:robust-sample}
\end{figure*}

\begin{figure*}[t]
    \centering
    \includegraphics[width=\linewidth]{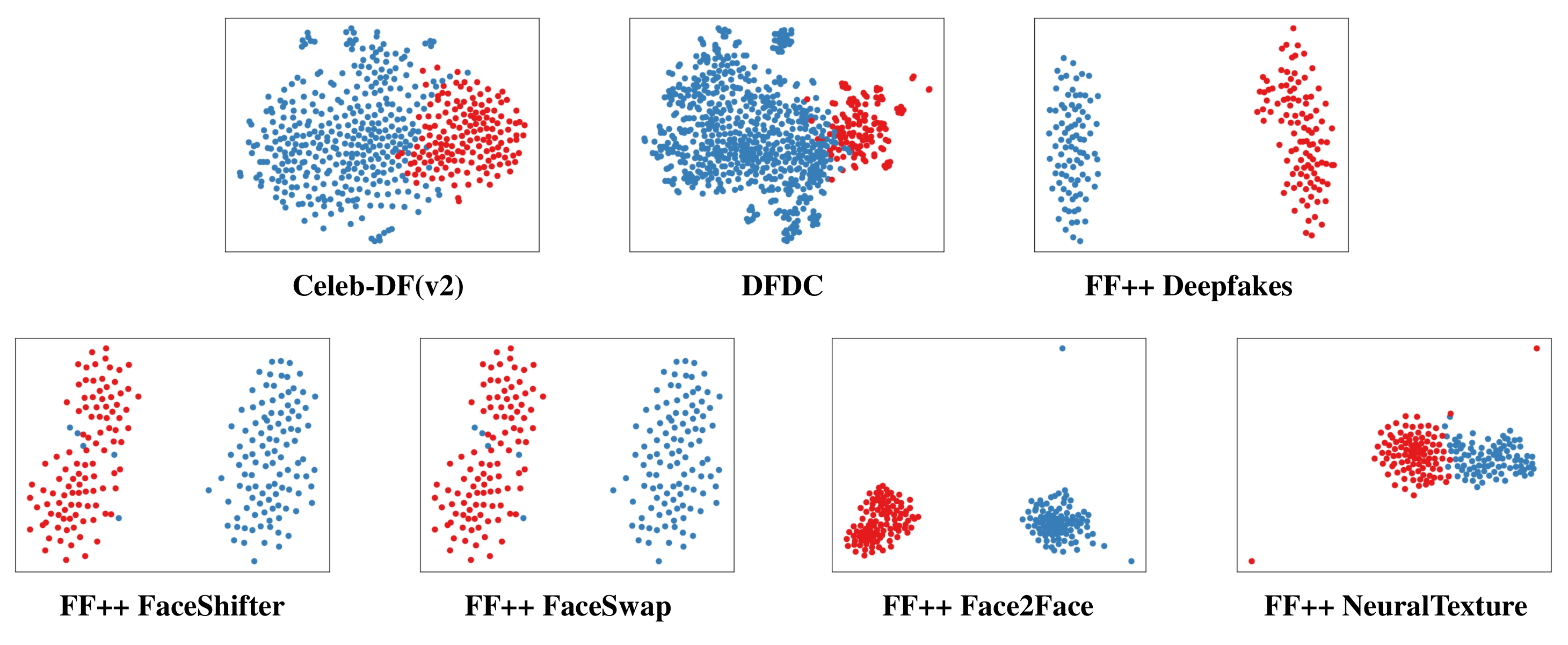}
    \caption{Visualization of representations on Celeb-DF(v2), DFDC and FF++~(five subsets) via t-SNE. Red: Real, Blue: Fake.}\label{figure:repre}
\end{figure*}

\subsection{Performance Analysis}

\begin{table}[ht]
   \centering
   \caption{Cross-datasets generalization results on Celeb-DF(v2) (C-DF), DFDC and FaceForensics++ Deepfakes~(FF-DF).}\label{tab:tab3-0}
       \begin{tabular}{ccccccc}
          \toprule
              {Train} & {Test} & {ACC(\%)} & {AUC(\%)} & {REC(\%)} & {PRE(\%)} & {F1(\%)}\\
              \midrule
              \multirow{2}{*}{C-DF} 
              & DFDC & 83.87 & 70.14 & 100.0 & 83.87 & 91.23\\
              & FF-DF & 74.62 & 79.26 & 87.10 & 62.79 & 72.97\\\midrule
               
              \multirow{2}{*}{DFDC} 
              & C-DF & 71.43 & 73.60 & 95.59 & 70.96 & 81.45\\
              & FF-DF & 80.71 & 87.95 & 92.47 & 69.36 & 79.26\\\midrule
               
              \multirow{2}{*}{FF-DF} 
              & C-DF & 68.92 & 69.78 & 96.47 & 68.76 & 80.29\\
              & DFDC & 83.87 & 66.99 & 100.0 & 83.87 & 91.23\\
          \bottomrule
       \end{tabular}
\end{table}

\myPara{Robustness.}~To evaluate the robustness of our proposed approach, we train our model on the original datasets and test them under several augmentation and perturbation methods. We choose five representative methods, including Flip, Blur, Bright, Compress, and Gaussian Noise. Specifically, we utilize 10x10 kernel size in Blur, set compression ratio to 4 in Compress~(where file storage becomes 1/4 after compression), and set the mean and variance of Gaussian Noise to 0.1 and 0.01. Besides, we train the models without any additional augmentation methods. The results are presented in Tab.~\ref{tab:tab6}.
From the table, we could observe that the performance under different augmentation methods is nearly the same as origin, such as the ACC and AUC under Flip and Compress. This provides clear evidence for the impressive robustness of our proposed approach on handling real-world scenarios such as compressed videos on the Internet. Two video examples and corresponding output predictions during testing are presented in Fig.~\ref{figure:robust-sample} for better comprehension.

\myPara{Cross-dataset Generalizability.} To further demonstrate our approach's generalizability, we perform cross-dataset experiments on three datasets by training on one dataset then testing on the other two. The results are presented in Tab.~\ref{tab:tab3-0}.
From the results, our approach exhibits competitive AUC scores compared to those opponents in Tab~\ref{tab:tab1}. For example,  the model training on DFDC and testing on Celeb-DF and FF++, and the model training on Celeb-DF and testing on DFDC both outperform plenty of approaches listed in Tab.~\ref{tab:tab1}. This implies that our approach achieves an impressive generalization ability by capturing the spatial and temporal inconsistency generally existing in deepfake videos instead of focusing on the specific generation pattern or artifact of certain deepfake methods.
Moreover, in some real-world scenarios where we may not access to the specific generated data, we can directly utilize the model trained on other accessible data to perform detection based on the proven cross-dataset generalizability. Besides, if we can collect a small amount of data, fine-tuning the trained model could achieve better performance.

\myPara{Representation.}~To further demonstrate the representation ability of our approach, we utilize t-SNE~\cite{tSNE} to visualize the representations learned from trained transformer encoder on three datasets and each subsets. The visualization results are presented in Fig.~\ref{figure:repre}, which shows that the real and fake videos generated by different methods are distinctly clustered in latent space, proving the strong representation ability of our approach.

\begin{table}[ht]
   \centering
   \caption{Ablation analysis on operation effectiveness on Celeb-DF(v2): no-dropout~(-), temporal dropout~(T), spatial dropout~(S), and spatiotemporal dropout~(S+T).}\label{tab:tab4}
       \begin{tabular}{ccccccc}
          \toprule
              {Dropout} & {ACC(\%)} & {AUC(\%)} & {REC(\%)} & {PRE(\%)} & {F1(\%)} \\
              \midrule
              {-}& 87.65 & 93.43 & 94.71 & 85.70 & 90.96\\
              {S}& 88.42 & 95.01 & 97.94 & 86.27 & 91.74\\
              {T}& 89.58 & 95.56 & 92.65 & 91.57 & 92.11\\
              {S+T}& 91.70 & 97.21 & 95.01 & 92.55 & 93.76\\
          \bottomrule
       \end{tabular}
\end{table}

\begin{table}[ht]
   \centering
   \caption{The AUC~(\%) comparisons with~\cite{DBLP:conf/ijcai/ZhangLL0G21} under different dropout operations and network architectures on Celeb-DF(v2), DFDC and FaceForensics++(FF++).}\label{tab:dropout}
       \begin{tabular}{ccccc}
          \toprule
              {Architecture} & {Dropout} & {Celeb-DF} & {DFDC} & {FF++} \\
              \midrule
              \multirow{2}{*}{3DCNN} 
              &TD~\cite{DBLP:conf/ijcai/ZhangLL0G21}  & 88.83 & 78.97 & 72.22\\
              &STD & 93.39 & 85.87 & 92.70\\\midrule
               
              \multirow{2}{*}{ViT} 
              &TD ~\cite{DBLP:conf/ijcai/ZhangLL0G21} & 95.56 & 97.39 & 97.41\\
              &STD & 97.21 & 99.14 & 99.76\\
          \bottomrule
       \end{tabular}
\end{table}

\subsection{Ablation Analysis}
To systemically evaluate the key components of our approach, we conduct ablation experiments on Celeb-DF(v2) dataset from three aspects and present a complete analysis in the following.

\myPara{Dropout Operation.}~To check the effect of dropout operation, we first investigate four variants:~1) no dropout (-), 2) spatial dropout (S), 3) temporal dropout (T), and 4) spatiotemporal dropout (S+T). Here, spatial dropout discards random patches in frames but does not discard frames, while temporal dropout just randomly drops some frames from extracted sequence. The results are presented in Tab.~\ref{tab:tab4}, and we can observe that the accuracy and AUC can be improved when applying spatial or temporal dropout. Specially, combining both spatial and temporal dropout achieves the best performance due to the consideration of spatiotemporal inconsistency, demonstrating the effectiveness of spatiotemporal dropout. 
To further verify that, we conduct an experimental comparison with TD-3DCNN~\cite{DBLP:conf/ijcai/ZhangLL0G21} by using different dropout operations and network architectures, and report the results in Tab~\ref{tab:dropout} where remarkable improvement is achieved by our spatiotemporal dropout.

\begin{figure}[ht]
    \centering
    \includegraphics[width=0.75\linewidth]{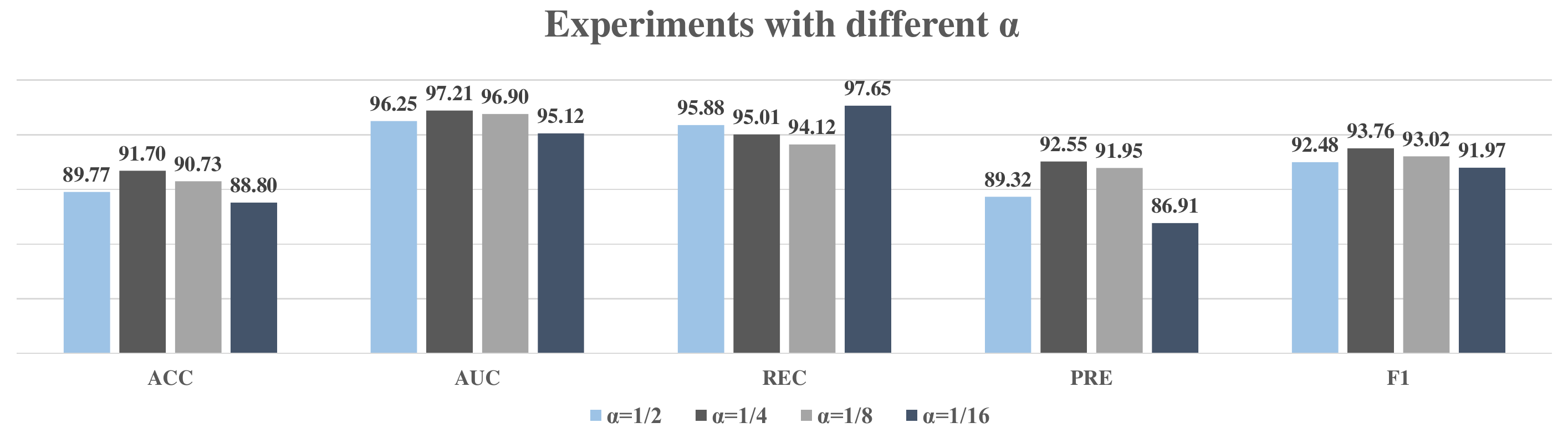}
    \caption{Ablation analysis on temporal dropout rate $\alpha$ on Celeb-DF(v2). We set $\alpha$ to 1/2, 1/4, 1/8, and 1/16 with other hyper-parameters fixed.}\label{figure:alpha}
\end{figure}

\begin{figure}[ht]
    \centering
    \includegraphics[width=0.75\linewidth]{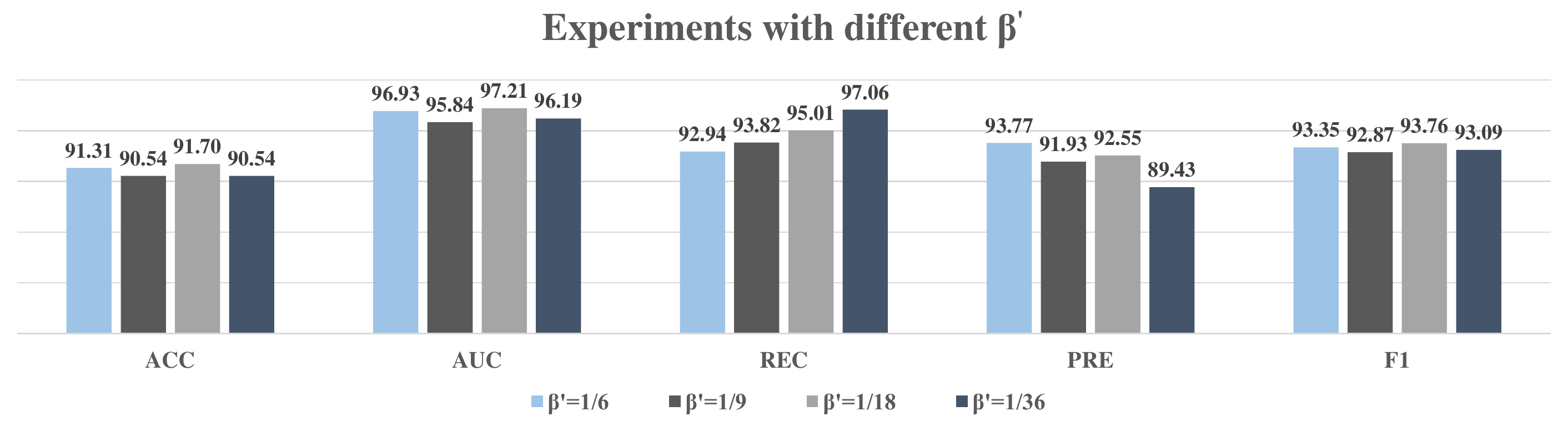}
    \caption{Ablation analysis on Spatial Dropout rate $\beta^{'} = 1 - \beta$ on Celeb-DF(v2). We set $\beta^{'}$ to 1/6, 1/9, 1/18, and 1/36 with other hyper-parameters fixed.}\label{figure:beta}
\end{figure}

\myPara{Dropout Rate.}~To further explore the effects of spatial and temporal dropout respectively, we evaluate our approach on Celeb-DF(v2) by setting different dropout rate $\alpha$ and $\beta$ with other hyper-parameters fixed. We define $\beta^{'} = 1 - \beta$ for convenience
and the results are presented in Fig.~\ref{figure:alpha} and Fig.~\ref{figure:beta}.
From the results, we can find that different $\alpha$ and $\beta$ result in different detection performance on accuracy and AUC, and the influence is significant, \textit{i.e.}, the AUC is about 2\% higher when $\alpha$ set to 1/4 compared to 1/16, and about 1.5\% higher when $\beta^{'}$ set to 1/18 compared to 1/9. Besides, we can find that the accuracy and AUC decline when setting $\alpha$ or $\beta$ too low or high,
\textit{i.e.}, the ACC and AUC both declined when changing $\alpha$ from 1/4 to 1/2 and 1/8, or changing $\beta^{'}$ from 1/18 to 1/9 and 1/36. 
We analyze this is because the spatiotemporal inconsistency is crushed or damaged when setting $\alpha$ or $\beta$ too low or high, dropping too many or few frames and patches. This indicates that the value of $\alpha$ and $\beta$ should be carefully considered.

\begin{table}[ht]
   \centering
   \caption{Ablation analysis of four different ViT backbones on Celeb-DF(v2) with ($\surd$) and without (-) STD.}\label{tab:tab5}
       \begin{tabular}{ccccccc}
          \toprule
              \multicolumn{1}{c}{Backbone} & \multicolumn{1}{c}{STD} & \multirow{1}{*}{ACC(\%)} & \multirow{1}{*}{AUC(\%)} & \multirow{1}{*}{REC(\%)} & \multirow{1}{*}{PRE(\%)} & \multirow{1}{*}{F1(\%)} \\
              \midrule
              \multirow{2}{*}{ViT-B16}
              & - & 82.43 & 92.63 & 79.41 & 92.79 & 85.58\\
              &$\surd$ & 91.70 & 97.21 & 95.01 & 92.55 & 93.76\\\midrule
               
               
              \multirow{2}{*}{ViT-B32}
              & - & 75.29 & 87.68 & 99.71 & 72.75 & 84.12\\
              &$\surd$ & 85.52 & 95.34 & 98.53 & 82.72 & 89.93\\\midrule
              
              \multirow{2}{*}{ViT-L16}
              & - & 76.83 & 89.27 & 97.06 & 75.01 & 84.62\\
              &$\surd$ & 86.87 & 94.26 & 91.47 & 88.57 & 90.14\\\midrule
              
              
              \multirow{2}{*}{ViT-L32}
              & - & 71.43 & 83.13 & 98.82 & 70.01 & 81.95\\
              &$\surd$ & 86.49 & 96.95 & 97.35 & 84.44 & 90.44\\
          \bottomrule
       \end{tabular}
\end{table}

\myPara{ViT Backbones.}~We evaluate our model's performance under four ViT backbones (ViT-B16, ViT-B32, ViT-L16 and ViT-L32~\cite{DBLP:conf/iclr/DosovitskiyB0WZ21}) on Celeb-DF(v2) to further demonstrate the effectiveness and flexibility of our spatiotemporal dropout. The results are presented in Tab.~\ref{tab:tab5}. From the table, we can find that the performance is consistently improved when incorporating STD, averagely improving ACC by 11.15\% and AUC by 7.76\%. This conclusively demonstrates the effectiveness of our STD as well as its flexibility to plug into various ViT architectures. Besides, the results shown in Tab.~\ref{tab:dropout} also come to this conclusion when employing in CNN architectures.

\section{Conclusion}
In this paper, we propose a spatiotemporal dropout transformer to detect deepfake videos at patch-level. In the approach, an input video is grid-wisely cropped and reorganized as massive \emph{bag of patches} instances which are then fed into a vision transformer to achieve robust representations. A spatiotemporal dropout operation is designed to fully explore the patch-level spatiotemporal inconsistency as well as serving as data augmentation, further improving model's robustness and generalizability. The spatiotemporal dropout operation is flexible and can be plugged into various ViTs. Extensive experiments clearly shows our approach outperforms 25 state-of-the-arts with impressive robustness, generalizability and representation ability. In the future, we will extend our approach to more video understand tasks and also enhance its interpretability to provide a more human-understandable detection result.



\bibliographystyle{unsrt}  
\bibliography{references}

\begin{thebibliography}{10}

\bibitem{DBLP:journals/tog/SuwajanakornSK17}
Supasorn Suwajanakorn, Steven~M. Seitz, and Ira Kemelmacher{-}Shlizerman.
\newblock Synthesizing obama: learning lip sync from audio.
\newblock {\em ACM TOG}, 36(4):95:1--13, 2017.

\bibitem{DBLP:journals/corr/abs-1912-13457}
Lingzhi Li, Jianmin Bao, Hao Yang, Dong Chen, and Fang Wen.
\newblock Faceshifter: Towards high fidelity and occlusion aware face swapping.
\newblock {\em arXiv preprint arXiv:1912.13457}, 2019.

\bibitem{lu2017attribute-guided}
Yongyi Lu, Yu{-}Wing Tai, and Chi{-}Keung Tang.
\newblock Attribute-guided face generation using conditional cyclegan.
\newblock In {\em ECCV}, pages 293--308, 2018.

\bibitem{thies2019deferred}
Justus Thies, Michael Zollh{\"{o}}fer, and Matthias Nie{\ss}ner.
\newblock Deferred neural rendering: image synthesis using neural textures.
\newblock {\em ACM TOG}, 38(4):66:1--12, 2019.

\bibitem{goodfellow2014generative}
Ian~J. Goodfellow, Jean Pouget{-}Abadie, Mehdi Mirza, Bing Xu, David
  Warde{-}Farley, Sherjil Ozair, Aaron~C. Courville, and Yoshua Bengio.
\newblock Generative adversarial nets.
\newblock In {\em NeurIPS}, pages 2672--2680, 2014.

\bibitem{DBLP:conf/iccp/PanZL12}
Xunyu Pan, Xing Zhang, and Siwei Lyu.
\newblock Exposing image splicing with inconsistent local noise variances.
\newblock In {\em ICCP}, pages 1--10, 2012.

\bibitem{DBLP:conf/sswmc/0002SS07}
Wen Chen, Yun~Q. Shi, and Wei Su.
\newblock Image splicing detection using 2-d phase congruency and statistical
  moments of characteristic function.
\newblock In {\em Security, Steganography, and Watermarking of Multimedia
  Contents IX}, volume 6505, page 65050R, 2007.

\bibitem{Rssler2019FaceForensics}
Andreas R{\"{o}}ssler, Davide Cozzolino, Luisa Verdoliva, Christian Riess,
  Justus Thies, and Matthias Nie{\ss}ner.
\newblock Faceforensics++: Learning to detect manipulated facial images.
\newblock In {\em ICCV}, pages 1--11, 2019.

\bibitem{li2020face}
Lingzhi Li, Jianmin Bao, Ting Zhang, Hao Yang, Dong Chen, Fang Wen, and Baining
  Guo.
\newblock Face x-ray for more general face forgery detection.
\newblock In {\em CVPR}, pages 5001--5010, 2020.

\bibitem{guera2018deepfake}
David Guera and Edward~J. Delp.
\newblock Deepfake video detection using recurrent neural networks.
\newblock In {\em AVSS}, pages 1--6, 2018.

\bibitem{DBLP:conf/ijcai/ZhangLL0G21}
Daichi Zhang, Chenyu Li, Fanzhao Lin, Dan Zeng, and Shiming Ge.
\newblock {Detecting Deepfake Videos with Temporal Dropout 3DCNN}.
\newblock In {\em IJCAI}, pages 1288--1294, 2021.

\bibitem{DBLP:conf/ih/AmeriniC20}
Irene Amerini and Roberto Caldelli.
\newblock Exploiting prediction error inconsistencies through lstm-based
  classifiers to detect deepfake videos.
\newblock In {\em IH{\&}MMSec}, pages 97--102, 2020.

\bibitem{DBLP:conf/icpr/TolosanaRFV20}
Rub{\'{e}}n Tolosana, Sergio Romero{-}Tapiador, Julian Fi{\'{e}}rrez, and
  Rub{\'{e}}n Vera{-}Rodr{\'{\i}}guez.
\newblock Deepfakes evolution: Analysis of facial regions and fake detection
  performance.
\newblock In {\em ICPR Workshops}, pages 442--456, 2020.

\bibitem{DBLP:conf/cvpr/0001VRTPT14}
Pablo Garrido, Levi Valgaerts, Ole Rehmsen, Thorsten Thorm{\"{a}}hlen, Patrick
  P{\'{e}}rez, and Christian Theobalt.
\newblock Automatic face reenactment.
\newblock In {\em CVPR}, pages 4217--4224, 2014.

\bibitem{DBLP:journals/tog/DaleSJVMP11}
Kevin Dale, Kalyan Sunkavalli, Micah~K. Johnson, et~al.
\newblock Video face replacement.
\newblock {\em ACM TOG}, 30(6):130, 2011.

\bibitem{thies2016face2face:}
Justus Thies, Michael Zollhofer, Marc Stamminger, Christian Theobalt, and
  Matthias Nie{\ss}ner.
\newblock Face2face: Real-time face capture and reenactment of rgb videos.
\newblock In {\em CVPR}, pages 2387--2395, 2016.

\bibitem{Li_2020_CVPR}
Lingzhi Li, Jianmin Bao, Hao Yang, Dong Chen, and Fang Wen.
\newblock Advancing high fidelity identity swapping for forgery detection.
\newblock In {\em CVPR}, pages 5074--5083, 2020.

\bibitem{DBLP:conf/cvpr/KarrasLA19}
Tero Karras, Samuli Laine, and Timo Aila.
\newblock A style-based generator architecture for generative adversarial
  networks.
\newblock In {\em CVPR}, pages 4401--4410, 2019.

\bibitem{DBLP:conf/cvpr/ChoiCKH0C18}
Yunjey Choi, Min{-}Je Choi, Munyoung Kim, Jung{-}Woo Ha, Sunghun Kim, and
  Jaegul Choo.
\newblock Stargan: Unified generative adversarial networks for multi-domain
  image-to-image translation.
\newblock In {\em CVPR}, pages 8789--8797, 2018.

\bibitem{DBLP:conf/cvpr/KarrasLAHLA20}
Tero Karras, Samuli Laine, Miika Aittala, Janne Hellsten, Jaakko Lehtinen, and
  Timo Aila.
\newblock Analyzing and improving the image quality of stylegan.
\newblock In {\em CVPR}, pages 8107--8116, 2020.

\bibitem{DBLP:conf/nips/VaswaniSPUJGKP17}
Ashish Vaswani, Noam Shazeer, Niki Parmar, Jakob Uszkoreit, Llion Jones, et~al.
\newblock Attention is all you need.
\newblock In {\em NeurIPS}, pages 5998--6008, 2017.

\bibitem{DBLP:conf/naacl/DevlinCLT19}
Jacob Devlin, Ming{-}Wei Chang, Kenton Lee, and Kristina Toutanova.
\newblock {BERT:} pre-training of deep bidirectional transformers for language
  understanding.
\newblock In {\em {NAACL-HLT}}, pages 4171--4186, 2019.

\bibitem{DBLP:conf/iclr/DosovitskiyB0WZ21}
Alexey Dosovitskiy, Lucas Beyer, Alexander Kolesnikov, Dirk Weissenborn,
  Xiaohua Zhai, et~al.
\newblock An image is worth 16x16 words: Transformers for image recognition at
  scale.
\newblock {\em arXiv preprint arXiv:2010.11929}, 2020.

\bibitem{carion2020end}
Nicolas Carion, Francisco Massa, Gabriel Synnaeve, Nicolas Usunier, Alexander
  Kirillov, and Sergey Zagoruyko.
\newblock End-to-end object detection with transformers.
\newblock In {\em ECCV}, pages 213--229, 2020.

\bibitem{DBLP:journals/corr/abs-2103-14030}
Ze~Liu, Yutong Lin, Yue Cao, Han Hu, Yixuan Wei, Zheng Zhang, Stephen Lin, and
  Baining Guo.
\newblock Swin transformer: Hierarchical vision transformer using shifted
  windows.
\newblock In {\em ICCV}, pages 10012--10022, 2021.

\bibitem{DBLP:journals/corr/abs-2103-15691}
Anurag Arnab, Mostafa Dehghani, Georg Heigold, Chen Sun, Mario Lu{\v{c}}i{\'c},
  and Cordelia Schmid.
\newblock Vivit: A video vision transformer.
\newblock In {\em ICCV}, pages 6836--6846, 2021.

\bibitem{DBLP:journals/corr/abs-2102-12122}
Wenhai Wang, Enze Xie, Xiang Li, Deng-Ping Fan, Kaitao Song, Ding Liang, Tong
  Lu, Ping Luo, and Ling Shao.
\newblock Pyramid vision transformer: A versatile backbone for dense prediction
  without convolutions.
\newblock In {\em ICCV}, pages 568--578, 2021.

\bibitem{han2020survey}
Kai Han, Yunhe Wang, Hanting Chen, Xinghao Chen, Jianyuan Guo, Zhenhua Liu,
  Yehui Tang, An~Xiao, Chunjing Xu, Yixing Xu, et~al.
\newblock A survey on visual transformer.
\newblock {\em arXiv preprint}, 2020.

\bibitem{DBLP:journals/corr/abs-2102-11126}
Deressa Wodajo and Solomon Atnafu.
\newblock Deepfake video detection using convolutional vision transformer.
\newblock {\em arXiv preprint arXiv:2102.11126}, 2021.

\bibitem{DBLP:journals/corr/abs-2107-02612}
Davide~Alessandro Coccomini, Nicola Messina, Claudio Gennaro, and Fabrizio
  Falchi.
\newblock Combining efficientnet and vision transformers for video deepfake
  detection.
\newblock In {\em ICIAP}, pages 219--229, 2022.

\bibitem{DBLP:journals/corr/abs-2104-01353}
Young~Jin Heo, Young~Ju Choi, Young{-}Woon Lee, and Byung{-}Gyu Kim.
\newblock Deepfake detection scheme based on vision transformer and
  distillation.
\newblock {\em arXiv preprint arXiv:2104.01353}, 2021.

\bibitem{dolhansky2019the}
Brian Dolhansky, Russ Howes, Ben Pflaum, Nicole Baram, and Cristian~Canton
  Ferrer.
\newblock The deepfake detection challenge (dfdc) preview dataset.
\newblock {\em arXiv preprint arXiv:1910.08854}, 2019.

\bibitem{DBLP:conf/cvpr/LiYSQL20}
Yuezun Li, Xin Yang, Pu~Sun, Honggang Qi, and Siwei Lyu.
\newblock Celeb-{DF}: {A} large-scale challenging dataset for deepfake
  forensics.
\newblock In {\em CVPR}, pages 3204--3213, 2020.

\bibitem{DBLP:conf/cvpr/ZhouHMD17}
Peng Zhou, Xintong Han, Vlad~I. Morariu, and Larry~S. Davis.
\newblock Two-stream neural networks for tampered face detection.
\newblock In {\em CVPRW}, pages 1831--1839, 2017.

\bibitem{DBLP:conf/wifs/AfcharNYE18}
Darius Afchar, Vincent Nozick, Junichi Yamagishi, and Isao Echizen.
\newblock Mesonet: a compact facial video forgery detection network.
\newblock In {\em WIFS}, pages 1--7, 2018.

\bibitem{yang2019exposing}
Xin Yang, Yuezun Li, and Siwei Lyu.
\newblock Exposing deep fakes using inconsistent head poses.
\newblock In {\em ICASSP}, pages 8261--8265, 2019.

\bibitem{8638330}
F.~{Matern}, C.~{Riess}, and M.~{Stamminger}.
\newblock Exploiting visual artifacts to expose deepfakes and face
  manipulations.
\newblock In {\em WACVW}, pages 83--92, 2019.

\bibitem{nguyen2019multi}
Huy~H Nguyen, Fuming Fang, Junichi Yamagishi, and Isao Echizen.
\newblock Multi-task learning for detecting and segmenting manipulated facial
  images and videos.
\newblock In {\em BTAS}, pages 1--8, 2019.

\bibitem{DBLP:conf/cvpr/LiL19c}
Yuezun Li and Siwei Lyu.
\newblock Exposing deepfake videos by detecting face warping artifacts.
\newblock In {\em CVPRW}, pages 46--52, 2019.

\bibitem{DBLP:journals/corr/abs-1910-12467}
Huy~H. Nguyen, Junichi Yamagishi, and Isao Echizen.
\newblock Use of a capsule network to detect fake images and videos.
\newblock {\em arXiv preprint arXiv:1910.12467}, 2019.

\bibitem{sabir2019recurrent}
Ekraam Sabir, Jiaxin Cheng, Ayush Jaiswal, Wael AbdAlmageed, Iacopo Masi, and
  Prem Natarajan.
\newblock Recurrent convolutional strategies for face manipulation detection in
  videos.
\newblock In {\em CVPRW}, pages 80--87, 2019.

\bibitem{DBLP:conf/cvpr/WangW0OE20}
Sheng-Yu Wang, Oliver Wang, Richard Zhang, Andrew Owens, and Alexei~A Efros.
\newblock {CNN}-generated images are surprisingly easy to spot... for now.
\newblock In {\em CVPR}, pages 8695--8704, 2020.

\bibitem{DBLP:conf/eccv/MasiKMGA20}
Iacopo Masi, Aditya Killekar, Royston~Marian Mascarenhas, Shenoy~Pratik
  Gurudatt, and Wael AbdAlmageed.
\newblock Two-branch recurrent network for isolating deepfakes in videos.
\newblock In {\em ECCV}, pages 667--684, 2020.

\bibitem{DBLP:conf/eccv/ChaiBLI20}
Lucy Chai, David Bau, Ser-Nam Lim, and Phillip Isola.
\newblock What makes fake images detectable? understanding properties that
  generalize.
\newblock In {\em ECCV}, pages 103--120, 2020.

\bibitem{mittal2020emotions}
Trisha Mittal, Uttaran Bhattacharya, Rohan Chandra, Aniket Bera, and Dinesh
  Manocha.
\newblock Emotions don't lie: An audio-visual deepfake detection method using
  affective cues.
\newblock In {\em ACM MM}, pages 2823--2832, 2020.

\bibitem{zhao2021multi}
Hanqing Zhao, Wenbo Zhou, Dongdong Chen, Tianyi Wei, Weiming Zhang, and Nenghai
  Yu.
\newblock Multi-attentional deepfake detection.
\newblock In {\em CVPR}, pages 2185--2194, 2021.

\bibitem{DBLP:conf/cvpr/HaliassosVPP21}
Alexandros Haliassos, Konstantinos Vougioukas, Stavros Petridis, and Maja
  Pantic.
\newblock Lips don't lie: A generalisable and robust approach to face forgery
  detection.
\newblock In {\em CVPR}, pages 5039--5049, 2021.

\bibitem{DBLP:conf/ijcai/HuXWLW021}
Ziheng Hu, Hongtao Xie, Yuxin Wang, Jiahong Li, Zhongyuan Wang, and Yongdong
  Zhang.
\newblock Dynamic inconsistency-aware deepfake video detection.
\newblock In {\em IJCAI}, pages 736--742, 2021.

\bibitem{liu2021spatial}
Honggu Liu, Xiaodan Li, Wenbo Zhou, Yuefeng Chen, Yuan He, Hui Xue, Weiming
  Zhang, and Nenghai Yu.
\newblock Spatial-phase shallow learning: rethinking face forgery detection in
  frequency domain.
\newblock In {\em CVPR}, pages 772--781, 2021.

\bibitem{DBLP:conf/cvpr/Zhu0FLL21}
Xiangyu Zhu, Hao Wang, Hongyan Fei, Zhen Lei, and Stan~Z Li.
\newblock Face forgery detection by 3d decomposition.
\newblock In {\em CVPR}, pages 2929--2939, 2021.

\bibitem{DBLP:journals/corr/abs-2203-02195}
Harry Cheng, Yangyang Guo, Tianyi Wang, Qi~Li, Tao Ye, and Liqiang Nie.
\newblock Voice-face homogeneity tells deepfake.
\newblock {\em arXiv preprint arXiv:2203.02195}, 2022.

\bibitem{DBLP:journals/corr/abs-2203-01318}
Xiaoyi Dong, Jianmin Bao, Dongdong Chen, Ting Zhang, Weiming Zhang, Nenghai Yu,
  Dong Chen, Fang Wen, and Baining Guo.
\newblock Protecting celebrities from deepfake with identity consistency
  transformer.
\newblock In {\em CVPR}, pages 9468--9478, 2022.

\bibitem{tSNE}
Van Der~Maaten Laurens and Geoffrey Hinton.
\newblock Visualizing data using t-sne.
\newblock {\em JMLR}, 9(2605):2579--2605, 2008.

\end{thebibliography}

\end{document}